\documentclass[10pt, conference]{IEEEtran}

\usepackage{diagbox}
\usepackage{graphicx,epsfig,cite}
\usepackage{comment}
\usepackage{caption}
\usepackage{subcaption}
\usepackage{lipsum}

\usepackage[letterpaper, left=0.55in, right=0.55in, bottom=0.73in, top=0.73in]{geometry}

\usepackage{amssymb,amsmath,amsfonts,dsfont,hyphenat}
\usepackage{fixltx2e,mathptmx}
\usepackage{latexsym,tabularx,comment,colortbl,epstopdf,xspace}
\usepackage{times,multirow,moreverb,booktabs}
\usepackage{accents,tikz,xfrac,xfrac}
\usepackage{paralist,fancyref,enumitem}

\newcommand{\minus}{\scalebox{0.75}[1.0]{$-$}}

\newcommand{\eat}[1]{}
\usepackage{color}

\usepackage{multirow}
\PassOptionsToPackage{bookmarks=false}{hyperref}

\IEEEoverridecommandlockouts

\begin{document}

\title{Frequency-based Multi Task learning With Attention Mechanism for Fault Detection In Power Systems}
\author{\large Peyman Tehrani and Marco Levorato\\
\normalsize Donald Bren School of Information and Computer Sciences, UC Irvine\\
\normalsize e-mail:\{peymant,~levorato\}@uci.edu\thanks{This work was supported in part by UCOP under Grant LFR-18-548175.}\vspace{-0.5cm}}
\date{}

\maketitle

\vspace{1cm}

\begin{abstract}

The prompt and accurate detection of faults and abnormalities in electric transmission lines is a critical challenge in smart grid systems. Existing methods mostly rely on model-based approaches, which may not capture all the aspects of these complex temporal series. Recently, the availability of data sets collected using advanced metering devices, such as Micro-Phasor Measurement units ($\mu$ PMU), which provide measurements at microsecond timescale, boosted the development of data-driven methodologies. In this paper, we introduce a novel deep learning-based approach for fault detection and test it on a real data set, namely, the Kaggle platform for a partial discharge detection task.  Our solution adopts a Long-Short Term Memory architecture with attention mechanism to extract time series features, and uses a 1D-Convolutional Neural Network structure to exploit frequency information of the signal for prediction. Additionally, we propose an unsupervised method to cluster signals based on their frequency components, and apply multi task learning on different clusters. The method we propose outperforms the winner solutions in the Kaggle competition and other state of the art methods in many performance metrics, and improves the interpretability of analysis.
\end{abstract}


\section{Introduction}
 

The increasing complexity and size of modern smart grid systems makes their monitoring and control more challenging.
These systems feature a combination of networking and sensing technologies, and computational subsystems to enable state estimation, event detection, and optimal control, where a large number of sensors are deployed to accurately and comprehensively monitor infrastructures and network status information, such as voltage, current, temperature, humidity, frequency, and so on\cite{dileep2020survey}.
 One of the most critical tasks, then, is to develop detection algorithms protecting the system against cyber-attacks or power-line faults. 
 To this aim, high-precision microphasor measurement units ($\mu$PMUs) \cite{pinte2015low}, provide high-fidelity voltage and current measurements to the monitoring system. The high resolution data from sensors has the potential to enable  advanced diagnostic and control applications. However, the high dimensionality of the data, and their complex patterns, make effective and efficient methodologies necessary to extract and analyze information from these signals.

Several approaches have been proposed for attack and fault detection in smart grids. Model-based methods has been employed in \cite{zhou2016abnormal,cordova2018shape,mivsak2016novel}. Most of them use PCA and SVD for features dimensionality reduction and to set the threshold used to detect anomalies. Importantly, most existing methods in this class ignore the temporal dependence that characterizes time series data. In \cite{hannon2019real}, authors proposed a vector autoregressive model (VAR) to learn the dynamics of the system toward prediction, and, detect anomalies based on the residual of prediction. Schemes based on parametric models are vulnerable to model mismatch, a characteristic that limits their applicability.

Non-parametric (model-free) techniques are data-driven, and inherently robust to data model mismatch. In \cite{shahsavari2019situational}, classical machine learning methods, such as support vector machine, are used for event detection based on a real-world dataset which  obtained from two micro-PMUs. The availability of large datasets in the power system community is enabling the training of increasingly sophisticate model-free machine learning-based algorithms. For instance, advanced methods such as generative adversarial networks and unsupervised learning methods are presented in \cite{aligholian2019event} and \cite{aligholian2019unsupervised} using real world data. In \cite{wang2018distributed}, a deep auto encoder is proposed to learn the feature distribution associated with normal data and detect anomalies. A multi-task logistic low-ranked dirty model (MT-LLRDM) is proposed in \cite{gilanifar2019multi} and validated for real-time PMU streams. The method utilizes the similarities in the fault data streams among multiple locations across a power distribution network in order to improve detection performance.
\IEEEpubidadjcol

In spite of the success of this class of approaches, there is still a room for considerable improvement. In this paper, we design a novel data-driven framework based on techniques that achieved state of the art performance in many domains, including natural language processing and computer vision, but has not been yet explored in the class of problems addressed herein. Specifically we combine together a bi-directional Long-Short Term Memory (Bi-LSTM) classifier with Attention Mechanism \cite{vaswani2017attention} in order to capture the faulty patterns in the time domain and a convolutional neural network (CNN) architecture which uses frequency domain features.
Additionally, we propose a frequency based clustering algorithm that can cluster signals based on their major frequency components. This clustering approach allow us to use multi task learning on different clusters in order to boost up the classifier performance. 

The proposed classifier is trained using a real world dataset, namely the VSB Power Line Fault Detection dataset, which was obtained using high frequency measuring devices (40 Mhz sampling rate) corresponding to the next generation of $\mu$PMUs\cite{hamacek2012detector,mivsak2019towards}. The signals in the dataset are high dimensional time series. Solutions for this kind of data have not been yet fully explored due to lower sampling rate of the current conventional measuring devices. Herein, we develop a framework for feature extraction both in time and frequency domain for dimensionality reduction.
We show that our proposed significantly outperforms the Kaggle winner solution in terms of F1 score (12\% increase), and  total accuracy (4\% increase), and has comparable performance in terms of Matthews correlation coefficient (MCC) and area under the curve (AUC).  

 The rest of the paper is organized as follows. In Section \ref{data}, we illustrate the dataset and introduce an algorithm to preprocess the signals, primarily to extract its peaks. In Section \ref{freq}, we analyze the data in the frequency domain and propose a clustering algorithm based on frequency components. In Section \ref{classifier}, we introduce and describe in detail the proposed predictive model architecture. Section \ref{eval} shows numerical results, and Section \ref{conc} concludes the paper.

\section{Dataset Description and Analysis} \label{data}
In this section, we discuss and analyze the data set,
\begin{figure}
\centering
        \includegraphics[width=.9\columnwidth]{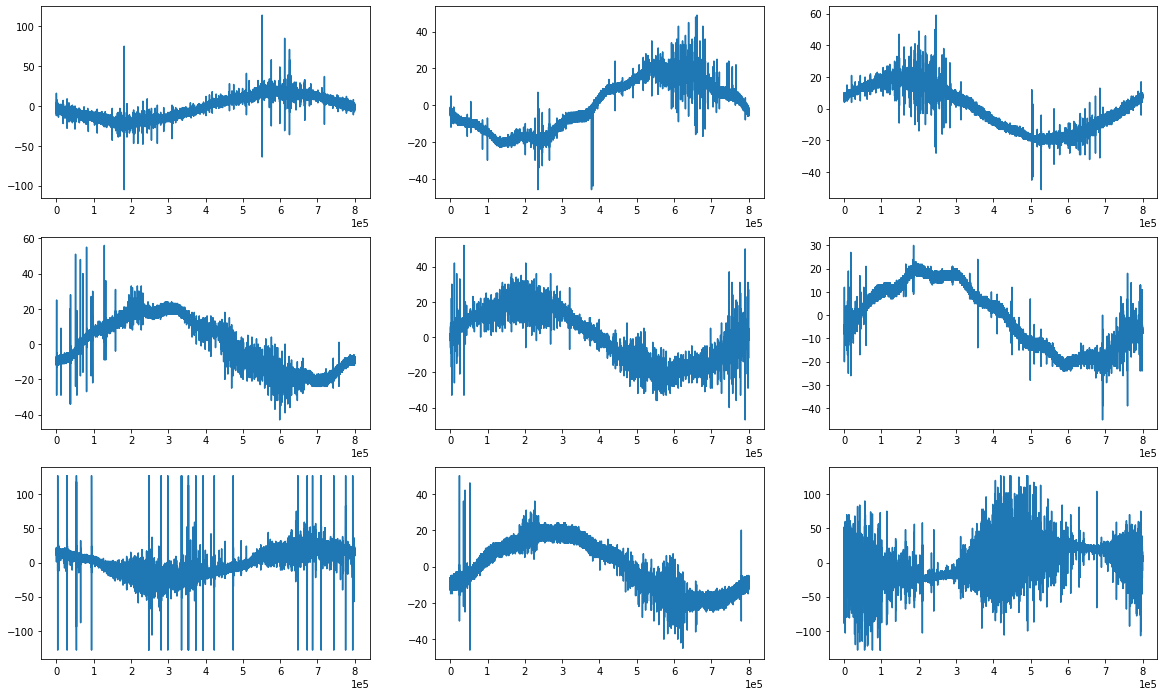}
        \caption{Signals with partial discharge (abnormal signals).}
        \label{fig:abnormal}
\end{figure}
which is publicly  available on the Kaggle website. The training and test datasets contains 8712 and 20337 datapoints, respectively, where each datapoint is a time series signal. Each signal contains 800,000 measurements of a power line's voltage, taken over 20 milliseconds. As the underlying electric grid operates at 50 Hz, this means that each signal encompasses a single complete grid cycle. The grid itself operates on a 3-phase power scheme, and all three phases are measured simultaneously. So in total we have 2904 and 6779 independent measurements for training and test data. To obtain this dataset, voltage signals  were measured with specific high frequency sampling devices \cite{hamacek2012detector} which can measure the power signal up to 40 MHz frequency, a relatively high frequency in this context. The measures were performed using metering devices at more than 20 different locations. As a consequence, the dataset incorporates a wide spectrum of noise and signal characteristics, which makes the design of an accurate and robust classifier even more challenging.

\begin{figure}
\centering
        \includegraphics[width=.9\columnwidth]{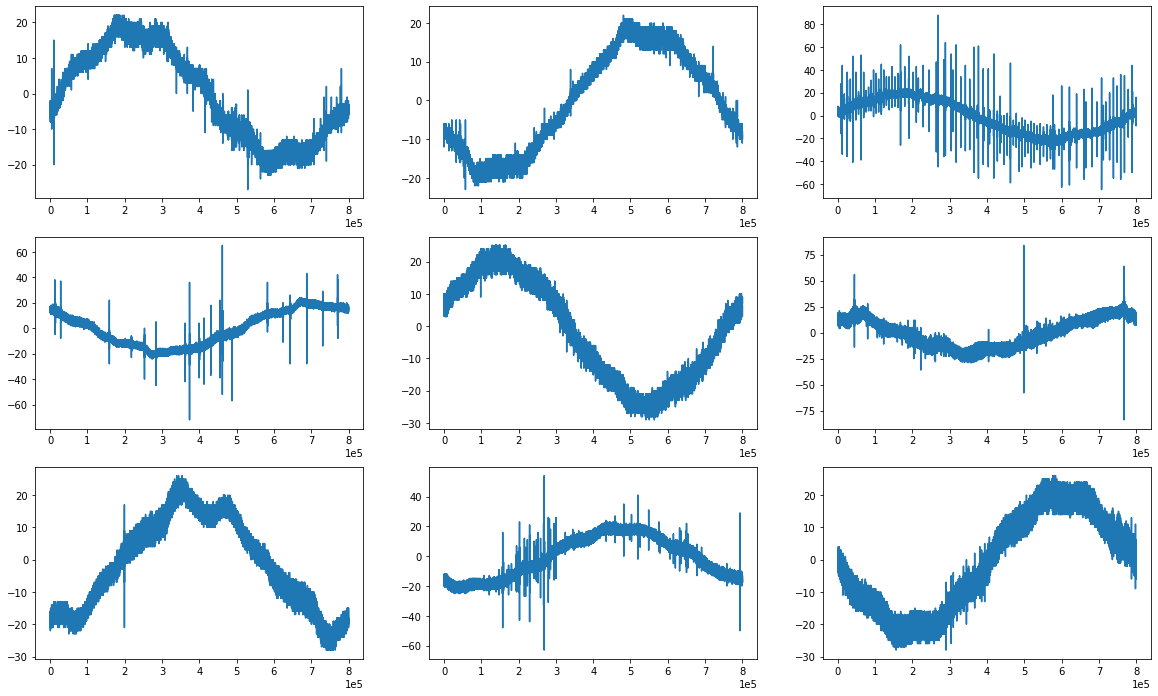}
        \caption{Normal signals.}
        \label{fig:normal}
\end{figure}

The labels in the data set are either zero or ones, where positive labels correspond to signals with partial discharge, and zero labels to normal signals. The distribution of labels is highly imbalanced, for instance, only 525 out of total 8712 are positive samples (less than $0.06$\%), which complicates the prediction problem. 
Fig.~\ref{fig:abnormal} and Fig.~\ref{fig:normal} show different samples of normal and partially discharged (PD) signals, respectively. It is apparent even from visual inspection how the number of peaks and their amplitude are important features that can contribute to discriminate the two classes. We also observe that even in the same class, the shape of signals can be quite different, another sign that the resulting classification problem is highly non-trivial.

\subsection{Time Domain Analysis}
We preprocess the signals to reduce the issues generated by the noise affecting them, as well as their high-dimensionality. We note that that each signal is composed of 800,000 samples. Transmitting this volume of data from each sensor to the cloud for analysis and decision making would consume a considerable amount of bandwidth and increase the probability of congestion, which in turn, would degrade performance in terms of detection latency.
The prepossessing algorithm is composed of 3 stages, high pass filtering, local maxima extraction, and then maxima sorting and thresholding. 
First, in order to reduce low frequency noises we apply to the signal a high-pass filter, which flattens the voltage signal and remove the signal phase, while preserving high frequency fluctuations and peaks in the signal.

\begin{figure}
\centering
        \includegraphics[width=\columnwidth]{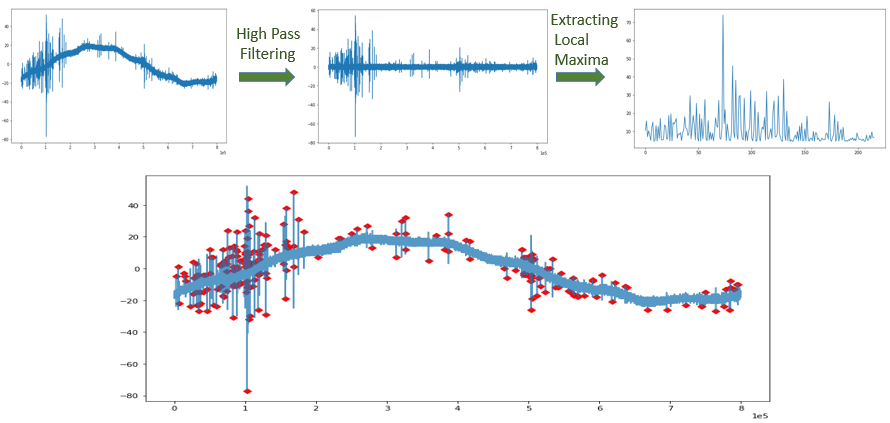}
        \caption{Peak extraction procedure.}
        \label{fig:peaks}
\end{figure}

After filtering, we pass the zero-phased signal to the peak detector algorithm which extracts the local maxima within a given neighborhood. We, then, sort the local maxima in descending order. When the difference between two consecutive sorted values is below the noise threshold we stop and eliminate the remaining smaller local maximas (noisy fluctuations) and keep the indexes of larger ones. Fig.~\ref{fig:peaks} shows the output signal after the high pass filtering and peak extraction procedures: the signal is flattened and low frequency components are removed after the high pass filtering, and only the main peaks of the original signal are extracted. It should be noticed that in addition to extracting the most informative parts of signal, the proposed algorithm reduces the signal size from 800,000 samples to hundreds of samples, a reduction of 3 orders of magnitude. The peak extraction procedure is described in the algorithm below.

\begin{table}[h!]
 \begin{tabular}{m{30em}}
 \hline
 \textbf{Peak Extraction Algorithm} \\
 \hline\hline
1.\text{Input} $X(n)$  \\ 
2. $Y=f(X) $ \quad \quad \quad \quad  High pass filtering \\
3.indx,Maximas= \text{Find Local maxima}(Y,neighbourhood) \quad \quad \quad \\
4.\text{Sort descending} (Maximas) \\
5. for $n=1$ to $N$ do \\
6.\quad if Maximas(n+1)-Maximas(n) $>$ threshold \\
7.  \quad save indx \\
8.\quad else \\
9.  \quad Stop \\
10.Return(X(indx))  \\
\hline
 \end{tabular}
\end{table}

\begin{figure}
\centering
        \includegraphics[width=\columnwidth]{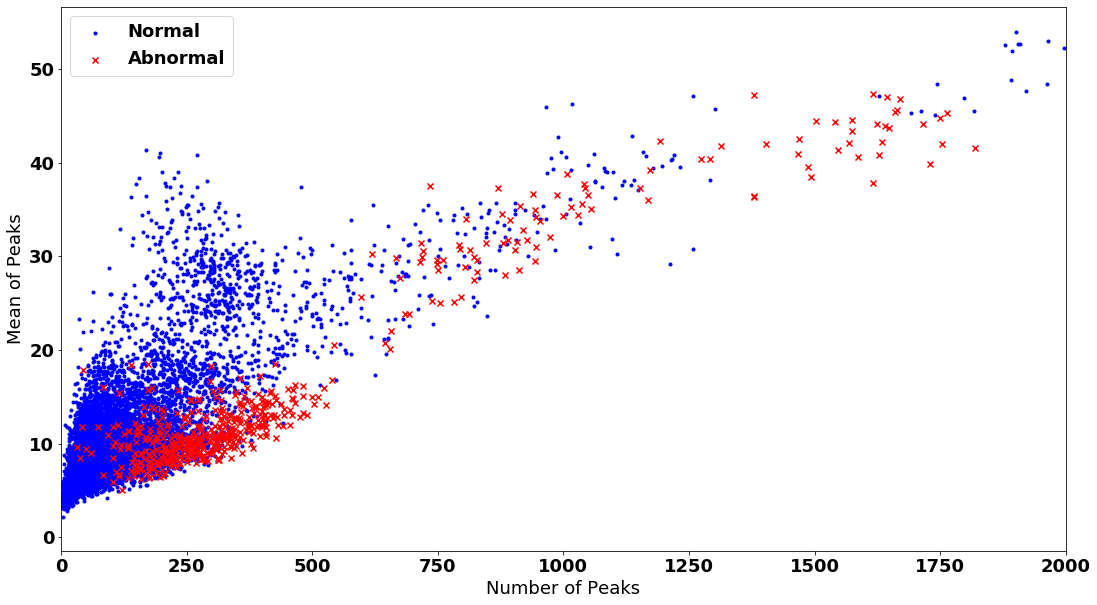}
        \caption{Scatter-plot of two classes based on Peak features.}
        \label{fig:scatter}
\end{figure}

Fig~\ref{fig:scatter} shows the separability of the signal based on the number of peaks and their mean values. It is evident that most of the normal signals' class has a smaller number of peaks, while many abnormal signals contain more than 200 peaks. This confirms that peaks contain a significant amount of information toward prediction.

\section{Frequency Based Feature Processing}\label{freq}
 One of the main challenges we face when dealing with real datasets, is that signals are often varied and their baseline features might be highly uncertain. In our context, several types of background noise and interference may affect the raw signal. As shown in Fig. \ref{fig:normal}, even normal signals may embed many  abrupt changes, transitions and patterns. In the problem at hand, there are several sources of background noise in medium voltage power lines \cite{mivsak2019towards}, such as discrete spectral interference, Repetitive pulse interference, random pulses interference, ambient and amplifier noise. As a consequence, the root cause of many peaks or high frequency patterns in the signal may not be necessarily related to partial discharge.
 
A thorough analysis of the signal in the frequency domain can help mitigating this issue, as the sources of interference and partial discharge patterns may express in different frequency components.
 To this end, one could simply use the discrete Fourier transform (DFT) coefficients of the signal computed using the FFT algorithm.
 We denote the DFT matrix as $W$, and define it as
\begin{equation}
W=
\begin{bmatrix}
1&1&1&\cdots&1 \\
1& w & w^2 & \cdots & w^{N-1} \\
1& w^2 & w^4 & \cdots & w^{2(N-1)} \\
\vdots&\vdots&\vdots&\vdots& \vdots \\
1& w^{N-1}&w^{2(N-1)}&\cdots&w^{(N-1)(N-1)} \\
\end{bmatrix},
\end{equation}
where $w=e^{\frac{-2\pi i }{N}}$ and  $X_w=WX$, in which $X_w$ is the signal representation in the frequency domain.
 The main issue of this approach is its high dimensionality, which could result in a very high number of coefficients, which contrast with the constrained computing power, memory and bandwidth available to the sensors capturing the signals. Even a simple classifier based on these features would be exceedingly complex.

A possible approach to reduce memory and complexity requirements is to choose a small subset of informative frequency components, effectively making the DFT matrix multiplication sparse. Principal component analysis (PCA) is one of the most popular algorithms to this end. The problem of using PCA is that it only captures the features with largest variation, which could not be necessarily informative about the class labels. Here, instead of PCA, we compute the mutual information (MI) as a measure of how much each DFT coefficient is informative and correlated with labels.
The mutual information between the frequency component and labels is defined as:
\begin{equation} \label{eu_eqn}
MI(X_w;Y)= \sum_{x_w,y}\log\frac{p(x_w,y)}{p(x_w)p(y)}
\end{equation}

\begin{figure}[t!]
\centering
        \includegraphics[width=.8\columnwidth]{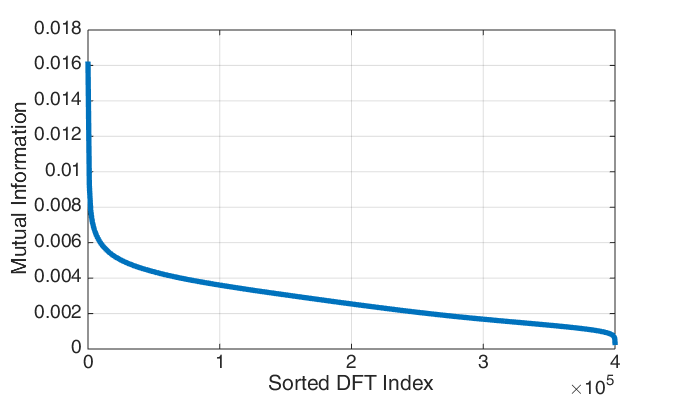}
        \caption{Sorted mutual information of frequency coefficients.}
        \label{fig:MI}
\end{figure}

\begin{figure}[h!]
\centering
        \includegraphics[width=.8\columnwidth]{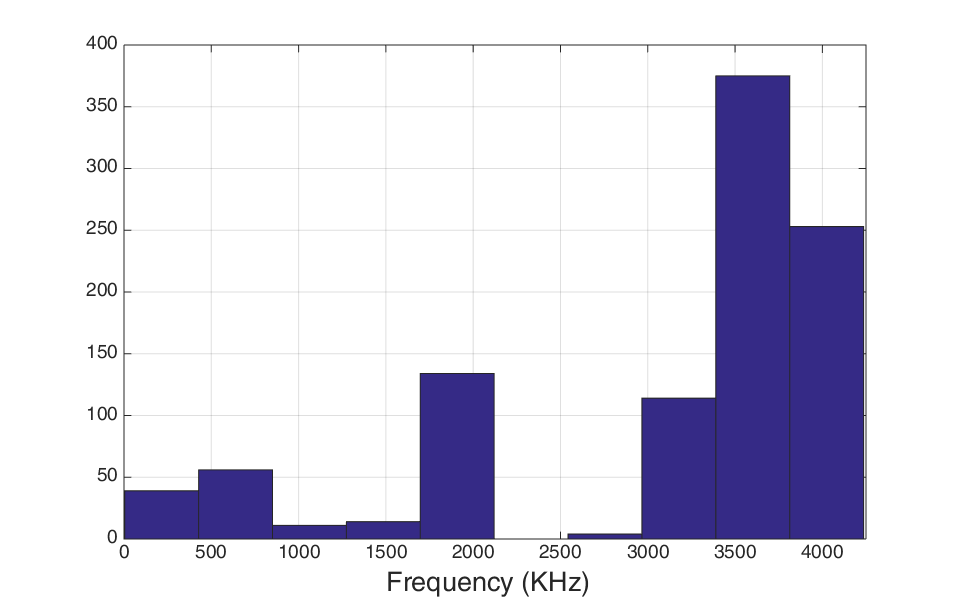}
        \caption{Histogram of top informative frequency bands.}
        \label{fig:histtop1}
\end{figure}

Fig.~\ref{fig:MI} shows the sorted frequency components based on their MI value. It can be seen how there is only a small set of coefficients which are highly informative toward the detection task. Fig. \ref{fig:histtop1} shows the histogram of the high informative frequency bands. Notably, the major frequency components of PD patterns are located between 3 Mhz to 4 Mhz. Also, some lower frequency bands could be useful toward the PD detection task.

Therefore, instead of all of the 8000000 rows of the matrix, we need only a small fraction of it, where we select only the rows associated with the most informative DFT coefficients (top $\%1$). One of the advantages of this method is that the mutual information of all the components need to be computed once in an offline manner, and then given the informative frequency indexes, only a sparse matrix multiplication needs to be implemented at the sensor level. The resulting computing task is extremely fast and efficient.

\subsection{Frequency Based Clustering}
\begin{figure}[t!]
\centering
        \includegraphics[width=.8\columnwidth]{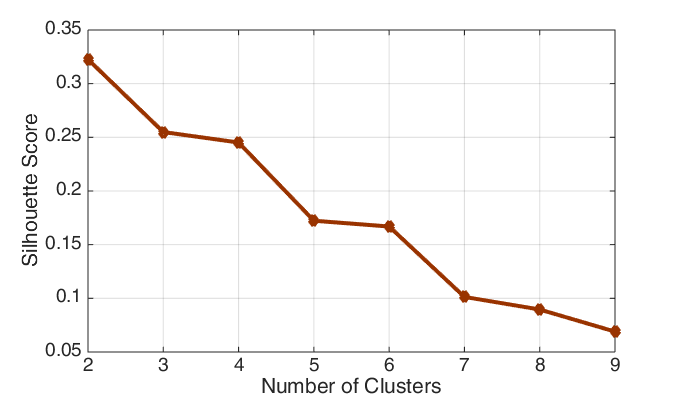}
        \caption{Silhouette clustering score vs number of clusters.}
        \label{fig:Cluster_score}
\end{figure}

Another interesting results of selecting only highly informative frequency coefficients is that signals can be grouped in an unsupervised manner into meaningful clusters. Fig~\ref{fig:Cluster_score} shows the Silhouette clustering score vs the number of clusters.
The Silhouette value $s(i)$, as defined below, is a measure of how similar an object is to its own cluster compared to other ones and it ranges from $\minus1$ to +1 , where a high value indicates that the object is better matched to its own cluster.
\begin{align} \label{eu_eqn}
s(i)=&\frac{b(i)-a(i)}{\max(a(i),b(i))}
\end{align}
where $a(i)$ and $b(i)$ are defined as:
\begin{align} 
a(i)=&\frac{1}{|C_i|-1}\sum_{j \in C_i,i\neq j}d(i,j) \\
b(i)=&\min_{k} \frac{1}{|C_k|}\sum_{j \in C_k}d(i,j), 
\end{align}
where $d(i,j)$ is the distance between two data points $i$ and $j$.
$a(i)$, then, is the mean distance between $i$ and all other data points in the same cluster, while and $b(i)$ is the smallest mean distance of $i$ to all points in other clusters.

\begin{figure}[h!]
\centering
        \includegraphics[width=.8\columnwidth]{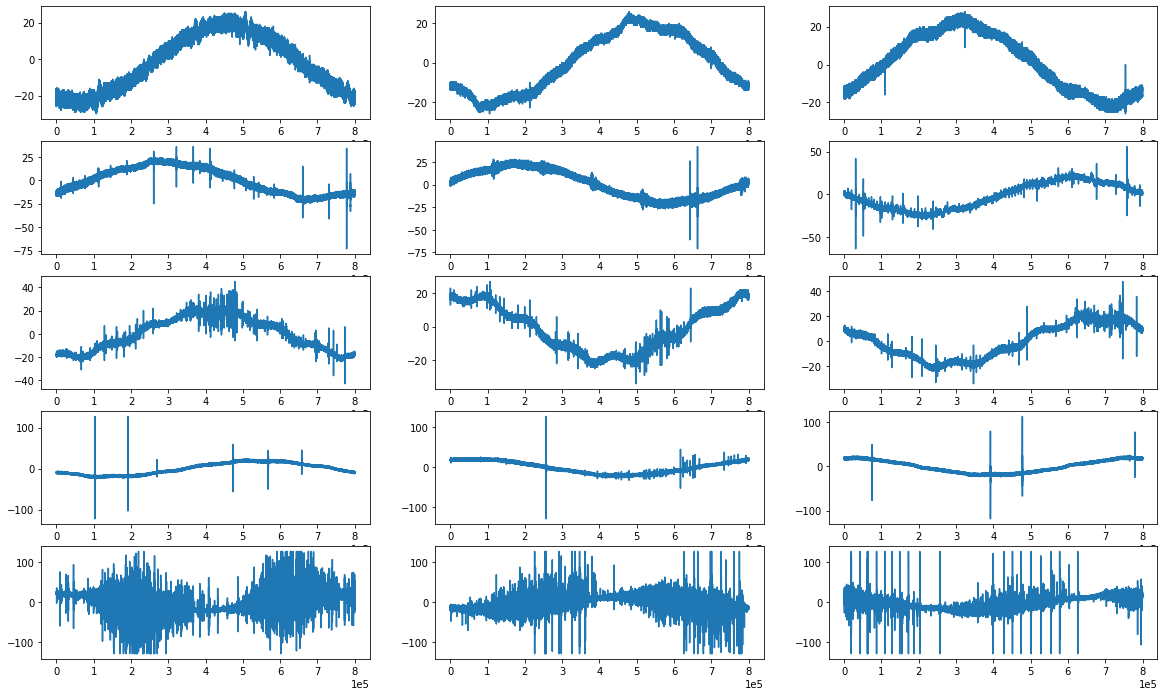}
        \caption{Clustering based on frequency components. Signals in the same row are at the same cluster.}
        \label{fig:signal_cluster}
\end{figure}

Here, we set the number of clusters to 5. Table \ref{tab:cluster} shows the total number of samples, the number and percentage of abnormal samples in each cluster. We can see that each cluster is an effective prior for the predictive model. For instance, in cluster 0 only 1\%  of the signals is abnormal, whereas $40$\% of cluster 4 is composed of abnormal signals. In Fig. \ref{fig:signal_cluster}, each row shows co-clustered signals. It is possible to see that clusters contain visually similar signals. 
These results point to using cluster assignment as an input to improve performance. In the proposed prediction model described in the next section,  we will use clustering to build a multi task learning approach, where we consider fault detection in each cluster as a separate task. This approach considerably boosts model performance.

\begin{table}[t!]
\begin{center}
\begin{tabular}{ |p{1cm}|p{2.3cm}|p{2.3cm}|p{1cm}|  }
 \hline
 Cluster & Num of Samples & Num of Pos Lables & Pos rate\\
 \hline
 0 & 5127 & 56  &   1.1 \% \\
  \hline

 1 & 1431 & 71  &  5.0 \% \\
  \hline

 2 & 994  & 124 &  12.5 \% \\
  \hline

 3 & 914  & 174 &  19.0 \% \\
  \hline

 4 & 246  & 100 & 40.7 \% \\

 \hline
\end{tabular}
\caption{\label{tab:cluster}Cluster details}
\end{center}
\end{table}

\section{Classifier}\label{classifier}
In this section, we describe  in detail the proposed classifier, which employs a deep learning architecture that takes both the time domain and frequency domain features for PD detection.

Recurrent Neural Networks (RNN) are a powerful class of neural networks designed to handle sequence dependence. Long Short Term Memory networks (LSTM) are a special kind of RNN specialized on the analysis of long-term dependencies \cite{hochreiter1997long}. 
One of the main issues with LSTMs is that this architecture may not be effective when the length of the sequence is too large, e.g., more than 1000 time samples. So in the problem at hand, in order to make the signal proper for using LSTMs, we  convert the time series input $X_n(T)$ into a multivariate time series with lower dimensions. Specifically, the original time series of length $T$ is divided into $m$ chunks of length $l$, where $T=m\times l$. For each chunk we compute $r$ different statistics of each chunk $\psi_r(x[m\times l:(m+1)\times l])$ (e.g., mean, median, mode, variance, different percentiles, skewness). The original signal $X_{1 \times T}$, then, is converted into the multivariate times series $X_{r \times m}^{chunks}$ represented by the matrix 

\begin{equation*}
\psi(X_{1\times T}) = X_{r\times m}^{chunks}=
\begin{pmatrix}
\psi_{1}(x_1) & \psi_{1}(x_2) & \cdots & \psi_r(x_m) \\
\psi_{2} (x_1)& \psi_{2}(x_2) & \cdots & \psi_r(x_m) \\
\vdots  & \vdots  & \ddots & \vdots  \\
\psi_{r} (x_1)& \psi_{r}(x_2) & \cdots & \psi_r(x_m)
\end{pmatrix}.
\end{equation*}

In order to make the LSTM classifier more effective, we adopt a Self Attention Mechanism, which has recently gained traction in many areas such as NLP, computer vision, medical signals analysis \cite{vaswani2017attention,feng2020spatio,cornia2018predicting}. Attention in deep learning can be broadly interpreted as a vector of importance weights. In order to predict or infer one element, such as a pixel in an image or a word in a sentence, using the attention vector we estimate how strongly it is correlated with other elements and take the sum of their values weighted by the attention vector as the approximation of the target. In the special case of partial discharge patterns, attention will give importance weights associated with the time regions of the signal which are more likely to be a PD pattern -- which mainly exist in positive labels. We remark that these regions cannot be effectively captured with conventional supervised learning methods or model based signal processing algorithms.

As we saw in the previous section, frequency domain analysis also provides us with valuable information toward classification. Therefore, in addition to the LSTM-Attention classifier -- which captures the PD patterns in the time domain -- we employ a CNN architecture  which utilizes the information in the frequency domain. CNNs are well known to fully exploit the topological dependencies which exist between the features and the class labels. The architecture is built using local connections and weights followed by pooling, which results in translation invariant features. As in our case, many DFT coefficients could be correlated based on their distance and the CNN operates as a feature extractor.

\begin{figure}
\centering
        \includegraphics[width=\columnwidth]{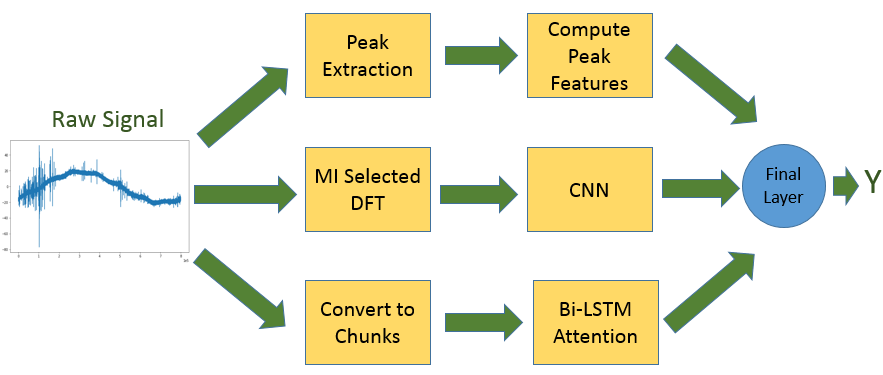}
        \caption{Architecture of the proposed prediction framework.}
        \label{fig:Classifier}
\end{figure}

Fig. \ref{fig:Classifier} show the proposed classifier architecture. Three types of features including: peak statistic, time series attention outputs and the CNN extracted frequency features  are fed into a final layer -- logistic regression -- to predict the class label. In order to avoid overfitting to the most frequent class, which often occurs  in imbalanced datasets, we used a weighted cross entropy loss function, instead of the naive cross entropy, as below:
\begin{align} 
J=\sum_{m=1}^M w_p y_1\log(h_{\theta}(x_m)+w_n (1-y_m)\log(1-h_{\theta}(x_m),
\end{align}
 where $w_p$ and $w_n$ are the weights that assigned to the positive and negative class, respectively. By setting the weight of each class to the inverse of its ratio to the whole data, the output of optimization is fair to all classes.

\section{Performance Evaluation}\label{eval}
In this section, we provide a thorough performance evaluation of the proposed classification algorithm. We implemented the model in PyTorch \cite{ketkar2017introduction} and used the Adam optimizer \cite{kingma2014adam} to train it with learning rate $lr=0.001$. We designed the CNN part with 6 hidden layers including two pooling and two 1-dimensional kernels with kernel sizes set to 5 and 10 respectively, and used Leaky ReLU  as activation function in the hidden layers. To form the matrix $ X_{r\times m}^{chunks}$ we used $r=19$ different statistics of $m=160$ chunks which will be fed into a two layer-stacked bidirectional LSTM followed by an Attention layer.

We briefly define two performance metrics that we will use in addition to obvious ones such as accuracy and area under the curve (AUC).


\vspace{1mm}
\noindent
{\bf F1-score} is a measure that combines precision and recall by taking the harmonic mean of these metrics:
$F_1=2 \times \frac{Precision \times Recall}{Precision + Recall}$

\vspace{1mm}
\noindent
{\bf The Matthews correlation coefficient (MCC)} takes into account the true and false positives and negatives and is generally regarded as a balanced measure, which can be used even if the classes are of very different sizes:
\begin{align} 
MCC=\frac{T_P \times TN -FP \times FN}{\sqrt{(TP+FP)(TP+FN)(TN+FP)(TN+FN)}}.
\end{align}


\vspace{1mm}
We compare our approach with several baseline classifiers, and in particular: 

\vspace{1mm}
\noindent
{\bf The Kaggel Winner} is a XGboost based classifier which uses the peaks features for prediction and achieved the first rank in the competition \cite{kagglefirst}. The authors trained more than 100 different  models on different random subsets of the train sets and used ensembling for final prediction. This strategy helps avoiding overfitting to the train-set since the training and test datasets have different distributions.

\vspace{1mm}
\noindent
{\bf LSTM with Attention} is the classifier which only uses the time domain PD patterns captured by the attention mechanism.

\vspace{1mm}
\noindent
{\bf CNN Freq} is a 1D-CNN architecture which uses only the frequency information as input features.

\vspace{1mm}
\noindent
{\bf DNN peaks} is a deep neural network which only uses the statistics of peaks as input features.

\vspace{1mm}

\begin{table}[!t]
\begin{center}

\begin{tabular}{ |p{2cm}|p{1cm}|p{1cm}|p{1.1cm}|p{1 cm}| }
 \hline
 Method & MCC & AUC & F1 Score & Accuracy \\
 \hline
 Kaggle Winner & 0.450 & 0.814  &   0.324 & 0.922  \\
  \hline

 LSTM-Attenstion & 0.387 & 0.797  &  0.29  & 0.913 \\
  \hline

 DNN with Peak features & 0.258  & 0.773 &  0.220  & 0.873 \\
  \hline

 CNN with frequency features & 0.293  & 0.779 &  0.272  & 0.917  \\
  \hline

 Proposed & 0.433  & 0.809 & \textbf{0.449}  & \textbf{0.961}  \\

 \hline
\end{tabular}
\caption{\label{tab:evaluation}Performance comparison between different methods}
\end{center}
\end{table}

Table \ref{tab:evaluation} compares the  performance measures achieved by different classifiers. The model we proposed is comparable to the Kaggle winner in terms of MCC and AUC (1-2\% difference),  while significantly outperforms it in terms of F1 score (more than 12\% improvement), and  total accuracy (around 4\% improvement). Note that unlike the Kaggle winner, we did not use ensembling by creating different models trained on a large number of subsample of train datasets, although this approach may improve performance. The motivation behind our choice is that our design realizes a structured and highly interpretable model, an important objective in this kind of application.

By comparing the proposed model with the  other three baselines, we emphasize how only by combining these feature (time, frequency and peaks) and employing frequency-based multi task learning obtain a very strong classifier.

\begin{figure}[!t]
\begin{subfigure}{.5\textwidth}
  \centering
  \includegraphics[width=\linewidth]{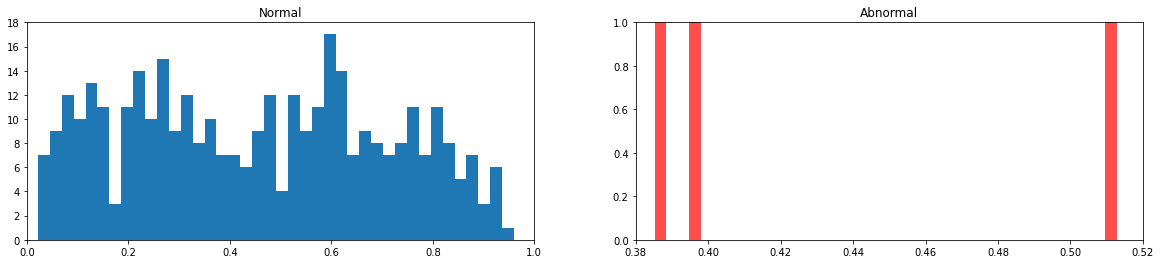}  
  \caption{Before Multi Task learning}
  \label{fig:sub-first}
\end{subfigure}
\begin{subfigure}{.5\textwidth}
  \centering
  \includegraphics[width=\linewidth]{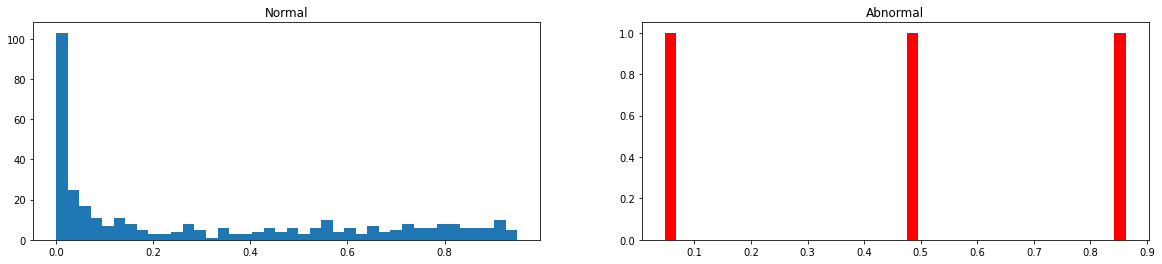}  
  \caption{After Multi Task Learning}
  \label{fig:sub-second}
\end{subfigure}
\caption{Effect of multi task learning on a frequency cluster.}
\label{fig:barcluster}
\end{figure}

Fig.~\ref{fig:barcluster} highlights the effect of multi task learning on one of the frequency clusters in the test dataset. The blue bars show the output probability of the model for normal signals and red ones are for the abnormal ones. After fine tuning the model weights on this cluster, the model learns that some patterns are not related to partially discharge, therefore setting the output probability of normal signals to be closer to zero, which allows the model to better differentiate between the two classes.

\begin{table}[!t]
\begin{center}

\begin{tabular}{ |p{2.2cm}|p{1.5cm}|p{1.5cm}|p{1.5cm}| }
 \hline
 Method & Peak Extraction & Statistics of Chunks & MI Selected DFT \\
 \hline
 Time Consumption &  5.58 (s) & 0.051 (s)  &   0.0002 (s)  \\

 \hline
\end{tabular}
\caption{\label{tab:time}Time consumption of different algorithms}
\end{center}
\end{table}
Tables \ref{tab:time} shows the time consumption for three different feature processing algorithms. Peak extraction has the largest delay since it includes filtering and lots of sorting in finding the local maximas. While converting the signal into the chunks and computing statistics of it  has a smaller delay comparing to peak extraction. Frequency feature processing is the fastest one as it only implies a sparse matrix multiplication and does not need any sorting, making it ideal for real-time applications.

 Inspired by our results and findings, we plan to consider a composite system where different sensors provide different types of feature available with heterogeneous delays and class discrimination capabilities. An optimization framework could be used to choose features in order to maximize an objective function capturing accuracy and delay measures.

\section{Conclusions}\label{conc}

In this paper, we proposed a classifier based on state of the art deep learning architectures, which uses both time domain and frequency domain features for fault and abnormality detection in smart grids. Additionally, we presented a frequency-based  clustering algorithm which can classify in an unsupervised fashion the signals into meaningful clusters. This allowed us to utilize multitask learning in our model. We demonstrated that our approach outperforms available baselines in most performance metrics.
\bibliographystyle{IEEEtran}
\bibliography{refnames}

\end{document}